\documentclass[lettersize,journal]{IEEEtran}
\usepackage{amsmath,amsfonts}
\usepackage{algorithmic}
\usepackage{algorithm}
\usepackage{array}
\usepackage[caption=false,font=normalsize,labelfont=sf,textfont=sf]{subfig}
\usepackage{textcomp}
\usepackage{stfloats}
\usepackage{url}
\usepackage{verbatim}
\usepackage{graphicx}
\usepackage{cite}
\usepackage{amsmath, amssymb}    
\usepackage{makecell}            
\usepackage{ulem}                
\usepackage{booktabs}           
\usepackage{array}               
\usepackage{xspace}               
\usepackage{soul}
\usepackage{wrapfig}
\usepackage{multirow}
\usepackage{makecell}

\usepackage[pagebackref,breaklinks,colorlinks,allcolors=teal]{hyperref}

\newcommand{\torecheck}[1]{#1}

\begin{document}

\title{Learning to Generate Cross-Task Unexploitable Examples}

\author{Haoxuan~Qu,
        Qiuchi~Xiang,
        Yujun~Cai,
        Yirui~Wu,
        Majid~Mirmehdi,
        Hossein~Rahmani,
        and~Jun~Liu
\thanks{H. Qu, Q. Xiang, H. Rahmani, and J. Liu are with Lancaster University, United Kingdom. Y. Cai is from The University of Queensland, Australia. Y. Wu is from Hohai University, China. M. Mirmehdi is from University of Bristol, United Kingdom.\\
E-mail: h.qu5@lancaster.ac.uk, q.xiang1@lancaster.ac.uk, yujun.cai@uq.edu.au, wuyirui@hhu.edu.cn, m.mirmehdi@bristol.ac.uk, h.rahmani@lancaster.ac.uk, j.liu81@lancaster.ac.uk}
\thanks{Manuscript received 
.\\(Haoxuan Qu and Qiuchi Xiang contributed equally to this work.) (Corresponding author: Jun Liu.)}}

\markboth{Journal of \LaTeX\ Class Files,~Vol.~14, No.~8, August~2021}%
{Shell \MakeLowercase{\textit{et al.}}: A Sample Article Using IEEEtran.cls for IEEE Journals}


\maketitle

\begin{abstract}
Unexploitable example generation aims to transform personal images into their unexploitable (unlearnable) versions before they are uploaded online, thereby preventing unauthorized exploitation of online personal images. Recently, this task has garnered significant research attention due to its critical relevance to personal data privacy. Yet, despite recent progress, existing methods for this task can still suffer from limited practical applicability, as they can fail to generate examples that are broadly unexploitable across different real-world computer vision tasks. To deal with this problem, in this work, we propose a novel \textbf{M}eta \textbf{C}ross-\textbf{T}ask \textbf{U}nexploitable \textbf{E}xample \textbf{G}eneration (\textbf{MCT-UEG}) framework. At the core of our framework, to optimize the unexploitable example generator for effectively producing broadly unexploitable examples, we design a \textit{flat-minima-oriented meta training and testing} scheme. Extensive experiments show the efficacy of our framework.
\end{abstract}


\definecolor{teal}{rgb}{0.21,0.49,0.74}

\section{Introduction}

\IEEEPARstart{O}{ver} the last decade, a large number of deep learning and computer vision models have been developed at an accelerated pace. Along this process, given that modern deep learning and computer vision models often require training on large-scale datasets to achieve optimal performance, deep learning practitioners are increasingly motivated to collect data from the Internet to improve their models \cite{radford2021learning,desai2021virtex}. However, this practice has raised significant concerns about the potential misuse of online data, particularly when it is gathered without the explicit consent of its owners. For instance, in 2020, a private company was found to have trained its commercial computer vision model using over 3 billion images scraped from the Internet without obtaining user consent \cite{hill2020secretive}. Such incidents have heightened public apprehension about the unauthorized, and even sometimes illegal, exploitation of their online personal data.

\begin{figure}[t]
\centering
\includegraphics[width=\columnwidth]{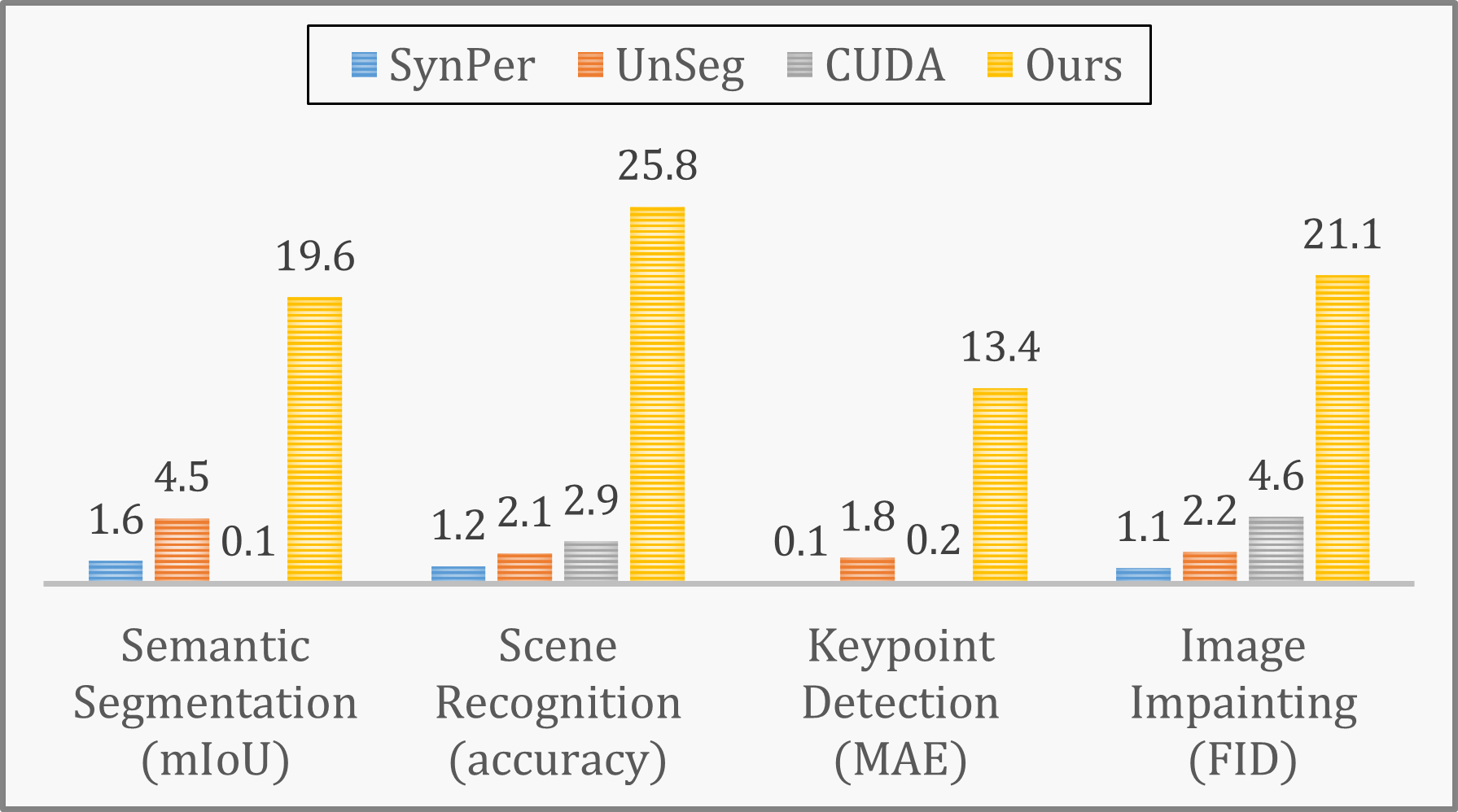}
\caption{Performance gap (drop) across various computer vision tasks when models for each task are trained on unexploitable examples compared to raw (clean) data. As shown, unexploitable examples generated by recent methods, including SynPer \cite{yu2022availability}, CUDA \cite{sadasivan2023cuda}, and UnSeg \cite{sun2024unseg}, fail to significantly and consistently reduce model performance. This implies that existing unexploitable example generation methods can fail to effectively prevent unauthorized exploitation across diverse real-world tasks, which can limit their practical utility. In contrast, our method is significantly more effective in generating cross-task unexploitable examples. All experiments are conducted on the Taskonomy dataset \cite{zamir2018taskonomy}, following Experimental Setting B.1 of Protocol B (details in Sec.~\ref{sec:experiments}). The evaluation metrics for each task are shown in brackets in the figure. (Best viewed in color.)} 
\label{fig:1}
\end{figure}

To deal with this problem, unexploitable example generation has attracted significant research attention \cite{huang2021unlearnable, fu2022robust, yu2022availability, sadasivan2023cuda, sun2024unseg}. This task typically focuses on transforming personal images into unexploitable (unlearnable) examples by adding imperceptible noise to each image (when it is uploaded online). This prevents deep learning models from learning useful information from online personal images during training, while still preserving the images' normal utility (e.g., social media sharing), as the noise added to each image is indiscernible to human eyes. 

Despite considerable progress, we contend here that existing unexploitable example generation methods can still struggle in real-world applications due to their limited cross-task generalization abilities. Specifically, in real-life scenarios, once a personal image is uploaded online, it can potentially be used as part of the training data for a wide variety of different computer vision tasks. Thus, for unexploitable examples to be truly useful in practice, they are naturally expected to exhibit \textit{cross-task unexploitable} properties, i.e., to \textit{broadly and robustly resist exploitation across different computer vision tasks with varying objectives}. Yet, existing unexploitable example generation methods \cite{huang2021unlearnable, fu2022robust, yu2022availability, sadasivan2023cuda, sun2024unseg} are typically designed to generate unexploitable examples for only a single task or a limited range of similar tasks. This can cause existing methods to struggle with effectively generating cross-task unexploitable examples (as shown in Fig.~\ref{fig:1}), significantly hindering their practical applicability.

Based on the discussion above, the goal of this paper is to develop a new unexploitable example generation framework that is more \textit{practically applicable}. Specifically, by optimizing the unexploitable example generator on only a limited set of available (seen) computer vision tasks, our framework aims to enable it to generate examples that are broadly unexploitable across both seen and unseen tasks in real-world scenarios. Achieving this goal, however, is non-trivial and ideally requires satisfying two key attributes, each posing its own challenges: 1) First, the unexploitable example generator needs to acquire knowledge that is genuinely relevant to \textit{cross-task generalizable unexploitability}. In practice, however, different computer vision tasks often have distinct objectives, meaning that perturbations effective for making images unexploitable in some tasks may not transfer to others \cite{sun2024unseg}. As such, how to guide the generator in our framework to effectively acquire cross-task unexploitable knowledge that generalizes even to unseen tasks after training on only a few seen tasks, presents a critical challenge. 2) Moreover, real-world scenarios often involve unforeseen variations, resulting in distribution shifts \cite{shift2008}. Thus, to generate broadly unexploitable examples effectively in practice, the generator typically also needs to robustly retain its learned (cross-task unexploitable) knowledge under distribution shifts. However, this robustness can be non-trivial to achieve in our context. As shown in \cite{yu2020gradient,chen2020just,phan2022improving}, incorporating multiple (seen) tasks into the optimization process can increase both the complexity and sharpness of the loss landscape. Consequently, without tailored guidance, the generator can be prone to converging toward sharp, rather than flat, minima in the loss landscape. Crucially, as illustrated in Fig.~\ref{fig:flat}, when the loss landscape exhibits a sharp minimum, even small distributional perturbations can result in significant increase in loss. This vulnerability, in turn, undermines the generator’s ability to robustly retain its learned (cross-task unexploitable) knowledge under distribution shifts. 

In this work, we handle both of the above challenges and propose an innovative framework for the practical and effective  generation of cross-task unexploitable examples,  refered to as \textbf{M}eta \textbf{C}ross-\textbf{T}ask \textbf{U}nexploitable \textbf{E}xample \textbf{G}eneration (\textbf{MCT-UEG}), from a novel perspective of meta-learning. Below, we outline our MCT-UEG framework in more detail.

Meta-learning, also known as ``\textit{learning to learn}'', enables virtual testing during model training to enhance the model's generalization ability \cite{finn2017model,nichol2018first}. 
Motivated by this, our framework carefully designs a \textit{flat-minima-oriented meta training and testing} scheme to optimize the generator and guide it to \textit{learn to effectively generate} broadly (generalizably) unexploitable examples.
Specifically, to perform this scheme, given a collection of (seen) computer vision tasks for training the generator, as a preparation, we first split these tasks into a meta-training set and a meta-testing set which contains different tasks with distinct objectives and characteristics. 

After that, using the constructed sets, our scheme operates in four steps: 
1) \textbf{Meta-training:} We begin by using tasks from the meta-training set to simulate a training step for the unexploitable example generator. 
2) \textbf{Meta-testing:} Next, the simulatedly-trained generator is evaluated on tasks in the meta-testing set, for its ability to generate unexploitable examples for tasks that hold distinct objectives compared to tasks in the meta-training set. This cross-task evaluation yields a feedback signal, the \textbf{meta-testing feedback}, that guides the generator in effectively acquiring cross-task unexploitable knowledge.
3) \textbf{Meta-flatness-assessment:} Concurrently with the meta-testing evaluation, we leverage gradient information from the simulated training step as a useful prior to also formulate a \textbf{meta-flatness feedback}. This feedback further guides the generator toward flat minima in the loss landscape, helping it robustly retain its learned unexploitable knowledge even under distribution shifts. Notably, this is achieved with minimal additional time cost, as our carefully designed meta-flatness feedback can be efficiently formulated largely in parallel with the meta-testing evaluation process (as discussed in Sec.~\ref{sec:meta}).
4) \textbf{Actual-optimization:} Finally, by jointly using the meta-testing feedback and the meta-flatness feedback, we adjust the simulated training step of the generator to actually optimize it. This joint feedback mechanism guides the generator to learn to simultaneously acquire cross-task unexploitable knowledge and robustly retain what it has learned, ultimately enhancing its ability to generate broadly unexploitable examples in a practical and effective manner.

The contributions of our work are as follows. 
1) We propose MCT-UEG, a novel framework for unexploitable example generation. To the best of our knowledge, MCT-UEG is the first approach to explore transforming personal images into broadly unexploitable examples, extending online personal image protection beyond a single task or a limited range of similar tasks. This enhances our framework’s potential in a wide range of real-world scenarios. 2) At the core of our framework, we design an innovative training scheme for the unexploitable example generator, facilitating its practically effective generation of cross-task unexploitable examples. 3) MCT-UEG achieves superior performance in unexploitable example generation across various tasks.

\section{Related Work}
\label{sec:related_work}

\noindent\textbf{Unexploitable Example Generation} has received considerable research attention 
\cite{huang2021unlearnable,fowl2021adversarial,yuan2021neural,fu2022robust,yu2022availability,sandoval2022autoregressive,sadasivan2023cuda,he2023indiscriminate,ren2023transferable,sun2024unseg,liu2024multimodal,zhao2024unlearnable,zhang2023unlearnable,ye2024ungeneralizable,fang2023towards,wang2024unlearnable,liu2024stable,chen2024one,ye2025how,yu2025mtlue,ma2025t2ue,li2025versatile} due to its practical significance.
In the early days, most methods for generating unexploitable examples, such as EM \cite{huang2021unlearnable}, TAP \cite{fowl2021adversarial}, NTGA \cite{yuan2021neural}, REM \cite{fu2022robust}, and MTL-UE \cite{yu2025mtlue}, required per-dataset optimization. Specifically, each time transforming a set of images into their unexploitable versions, even during inference, these methods typically require re-optimizing a neural network \cite{sadasivan2023cuda}. This requirement can hinder the usage of these methods in many real-world scenarios. More recently, several methods that are free from per-dataset optimization have been proposed. 
Yu et al. \cite{yu2022availability} proposed to first craft synthetic shortcuts using the method in \cite{guyon2003design} and then generate unexploitable examples by leveraging their crafted shortcuts. 
Sadasivan et al. \cite{sadasivan2023cuda} proposed generating unexploitable examples by leveraging randomly generated convolutional filters to perform controlled class-wise convolutions. 
Ma et al. \cite{ma2025t2ue} focused on scenarios where each personal image is paired with a corresponding (high-quality) text description, and proposed a text-guided approach that relies on image–text alignment to generate examples unexploitable across several image–text alignment tasks (e.g., text-to-image retrieval).
In addition, Sun et al. \cite{sun2024unseg} introduced UnSeg, which formulates the optimization of the unexploitable example generator as a bilevel min–min optimization problem based on interactive segmentation.

Existing unexploitable example generation methods are typically tailored for a single computer vision task \cite{huang2021unlearnable,fu2022robust,sadasivan2023cuda} or a limited set of similar tasks \cite{sun2024unseg,yu2025mtlue,ma2025t2ue}. Differently, in this work, we argue that enhancing the practicality of unexploitable example generation requires producing examples that are broadly unexploitable across a wide variety of different tasks, even including those previously unseen ones. To this end, we propose a novel framework that effectively generates such broadly unexploitable examples.

\noindent\textbf{Meta-Learning} \cite{finn2017model,nichol2018first,abbas2022sharp} was originally introduced in few-shot learning, where it has shown impressive capabilities. Since then, the idea of meta-learning has been also studied in various other tasks \cite{li2018learning,guo2020learning,huang2021metasets,qu2023towards,peng2024joint,meerza2024harmonycloak,liu2024metacloak}, such as face recognition \cite{guo2020learning}, point cloud classification \cite{huang2021metasets}, and action recognition \cite{peng2024joint}. Different from existing studies, in this work, we design an innovative \textit{flat-minima-oriented meta training and testing} scheme, which facilitates personal images in being effectively transformed into cross-task unexploitable examples.

\section{Proposed Method}

Given a personal image, the goal of our work is to generate its \textit{broadly unexploitable} version that preserves its unexploitable nature across a wide variety of computer vision tasks, both seen and unseen.
To achieve this, we propose a novel framework, MCT-UEG. 
Specifically, our framework is built upon the basic pipeline for training unexploitable example generators (as elaborated in Sec.~\ref{sec:base}), which has been successfully employed in recent methods such as \cite{sun2024unseg}. 
Our framework then introduces a \textit{flat-minima-oriented meta training and testing} scheme (as discussed in Sec.~\ref{sec:meta}), designed to help the generator learn to effectively and practically generate broadly unexploitable examples.

\subsection{Basic Training Pipeline}
\label{sec:base}

\torecheck{Following \cite{sun2024unseg}, we train the unexploitable example generator within our framework using an alternative optimization algorithm as the basic training pipeline.} In particular, let $T$ represent the \textit{limited} number of computer vision tasks available for training the unexploitable example generator (i.e., seen tasks). This basic training pipeline then involves the unexploitable example generator $f_g$ alongside a collection of $T$ surrogate models $\{f^t_s\}^T_{t=1}$, each corresponding to one of the available computer vision tasks. These surrogate models can either share parameters or hold their own parameters. Besides, both $f_g$ and $\{f^t_s\}^T_{t=1}$ are implemented as learnable neural networks. Below, we introduce in detail how $f_g$ is trained within this basic pipeline with the help of $\{f^t_s\}^T_{t=1}$.

Specifically, during training $f_g$ within this pipeline, given an input image $x$, following \cite{sun2024unseg}, we first pass $x$ into $f_g$ to derive its unexploitable version $x^u$ as:
\begin{equation}\label{eq:basic_1}
\begin{aligned}
x^u = \mathit{tanh}(f_g(x)) \times \epsilon + x,
\end{aligned}
\end{equation}
where $\mathit{tanh}(\cdot)$ represents the hyperbolic tangent function used to scale values in $f_g(x)$ to the range $[-1, 1]$, and $\epsilon$ as a hyperparameter represents the maximum allowable magnitude of noise that can be added to the input image $x$ to keep the difference between $x$ and $x^u$ still imperceptible. 
Next, after deriving $x^u$, to measure its unexploitable nature and thus enable the optimization of $f_g$, during each iteration of the optimization process, we further randomly select a surrogate model $f^t_s$ from the collection $\{f^t_s\}^T_{t=1}$, send $x^u$ into $f^t_s$, and calculate the loss $L$ based on $f^t_s(x^u)$ as:
\begin{equation}\label{eq:basic_2}
\begin{aligned}
L = L^t_s(f^t_s(x^u)),
\end{aligned}
\end{equation}
where $L^t_s$ is the loss function associated with $f^t_s$ (e.g., the cross-entropy loss function when $f^t_s$ is an image classification model). 
Our rationale here for calculating $L$ stems from prior analyses \cite{huang2021unlearnable,sun2024unseg}.
As analyzed in these works, a small loss $L$ for $f^t_s(x^u)$ indicates that the surrogate model $f^t_s$, after being trained on $x^u$, is insufficiently penalized by its objective function and can thus only be slightly updated, causing $f^t_s$ to fail to learn much useful information from $x^u$. This makes $L$ a reliable indicator of the unexploitable nature of $x^u$, allowing us to enhance the unexploitable nature of $x^u$ by minimizing $L$. Leveraging this insight, we get the overall objective of the basic training pipeline by combining Eq.~\ref{eq:basic_1} with Eq.~\ref{eq:basic_2} as:
\begin{equation}\label{eq:basic_3}
\begin{aligned}
\mathop{min} L, \text{where}~L = L^t_s\Big(f^t_s\big(\mathit{tanh}(f_g(x)) \times \epsilon + x\big)\Big).
\end{aligned}
\end{equation}
Building upon Eq.~\ref{eq:basic_3} as its objective, the basic training pipeline can finally be structured as a competitive game \cite{sun2024unseg} between $f_g$ and $\{f^t_s\}^T_{t=1}$. In specific, in the game, the pipeline alternates between updating $f_g$ (via $\mathop{min}\limits_{f_g} L$) and $f^t_s$ (via $\mathop{min}\limits_{f^t_s} L$) in a competitive manner.  
This iterative competition then enables the capabilities of $f_g$ (as well as $\{f^t_s\}^T_{t=1}$) to be gradually enhanced.

\subsection{Flat-Minima-Oriented Meta Training and Testing Scheme}
\label{sec:meta}

Above, we introduce a basic pipeline for training the unexploitable example generator. However, the efficacy of this pipeline, when applied naively by itself, remains suboptimal.
Specifically, as shown in Fig.~\ref{fig:1}, the recent method UnSeg \cite{sun2024unseg} that directly employs the above pipeline to train the unexploitable example generator, among other existing unexploitable example generation methods \cite{yu2022availability,sadasivan2023cuda}, can still fail to transform images into examples that are broadly unexploitable across different computer vision tasks. This lack of cross-task generalizability can significantly undermine the practical effectiveness of these methods, as personal images shared online, in practice, are at risk of being misused as training data for a wide variety of different computer vision tasks.

To handle this problem, we aim to empower the unexploitable example generator with the ability to produce broadly unexploitable examples, even for previously unseen computer vision tasks, in a practically effective manner. To achieve this, we first identify two key attributes that the generator’s training process should ideally satisfy. \ul{Attribute 1}: To facilitate the generator's effective generation of broadly unexploitable examples, on the one hand, its training process should be able to guide it in learning cross-task unexploitable knowledge. \ul{Attribute 2}: Meanwhile, to further enhance the generator's practical effectiveness, especially given the unforeseen distribution shifts commonly encountered in real-world scenarios, the generator's training process ideally should also encourage it to converge to flat minima in the loss landscape. As illustrated in Fig.~\ref{fig:flat}, this helps the generator robustly retain its learned (cross-task unexploitable) knowledge even under training-to-testing distribution shifts. To jointly satisfy these two attributes, our framework integrates a carefully designed \textit{flat-minima-oriented meta training and testing} scheme into the training pipeline introduced in Sec.~\ref{sec:base}. We describe this scheme in detail below and provide an illustration in Fig.~\ref{fig:2}.

\begin{figure}[t]
\centering
\includegraphics[width=\columnwidth]{Figures/main_paper_figure2.png}
\caption{Illustration of how flat minima help retain unexploitable knowledge under distribution shifts. We use the dashed green line to represent the training loss landscape and the solid green line to represent the testing loss landscape. As shown, when distribution shifts occur from training to testing, models converging to flat minima exhibit smaller increases in loss compared to those at sharp minima (as indicated by the red arrows). This suggests that flat minima confer greater robustness, enabling an unexploitable example generator converging to such minima to better preserve its learned unexploitable knowledge despite distribution shifts.  (Best viewed in color.)}
\label{fig:flat}
\end{figure}

As a preparatory step in our scheme, at the start of each epoch in which $f_g$ (instead of $\{f^t_s\}^T_{t=1}$) is to be optimized, we randomly split the $T$ seen computer vision tasks along with their corresponding surrogate models $\{f^t_s\}^T_{t=1}$ equally into two disjoint sets: the meta-training set $D^{m\_tr}$ and the meta-testing set $D^{m\_te}$. Constructed in such a mutually exclusive manner, these two sets then comprise surrogate models corresponding to tasks holding \textit{distinct objectives}. After this preparation, each iteration of training $f_g$ then proceeds through four steps: \textbf{meta-training}, \textbf{meta-testing}, \textbf{meta-flatness-assessment}, and \textbf{actual-optimization}. 

In particular, we first use the surrogate model in $D^{m\_tr}$ to simulate a training step for the generator $f_g$ (\textbf{meta-training}). We then evaluate the resulting simulatedly-trained generator using the surrogate model in $D^{m\_te}$ (\textbf{meta-testing}). Notably, the surrogate models here in $D^{m\_te}$ have been ensured to correspond to different tasks compared to the surrogate models in $D^{m\_tr}$. From this, we can intuit that, if the generator performs well during meta-testing after its simulated training in the meta-training step, it suggests that the generator has acquired knowledge relevant to cross-task unexploitability. This makes the feedback obtained from meta-testing, i.e., the \textbf{meta-testing feedback}, an effective supervisory signal for guiding the training pipeline to satisfy \ul{Attribute 1}. The remaining challenge is to further regularize the generator so that its training also meets \ul{Attribute 2}. 

\definecolor{myred}{RGB}{234,0,0}
\definecolor{myblue}{RGB}{100,162,222}
\definecolor{mygreen}{RGB}{76,163,45}
\definecolor{myyellow}{RGB}{253,189,7}

\begin{figure*}[t]
\centering
\includegraphics[width=0.88\textwidth]{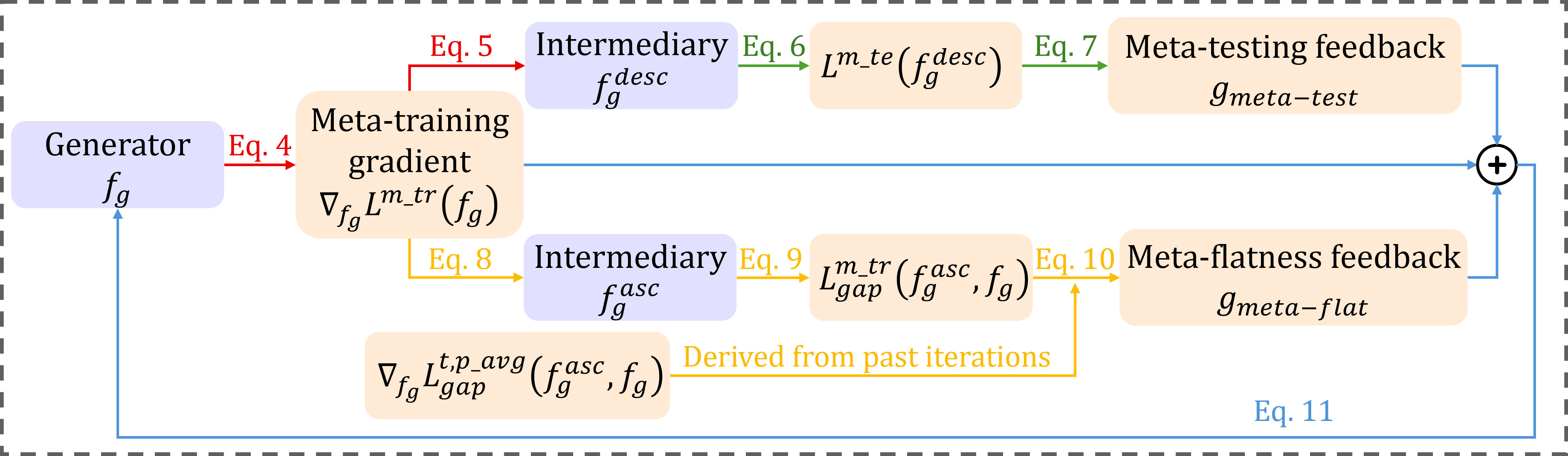}
\caption{Illustration of our \textit{flat-minima-oriented meta training and testing} scheme. This scheme consists of four steps: meta-training (indicated by \textcolor{myred}{red} arrows), meta-testing (\textcolor{mygreen}{green} arrows), meta-flatness-assessment (\textcolor{myyellow}{yellow} arrows), and actual-optimization (\textcolor{myblue}{blue} arrows). (Best viewed in color.)}
\label{fig:2}
\end{figure*}

To meet \ul{Attribute 2}, we first observe that the gradient derived from the simulated training step actually already encodes informative prior knowledge about the local flatness of the loss landscape around $f_g$'s current parameters. Leveraging this insight, in parallel to meta-testing, we also leverage this gradient prior to formulate a \textbf{meta-flatness feedback} (\textbf{meta-flatness-assessment}). This feedback helps regularize the generator toward flat minima in the loss landscape (as shown in Fig.~\ref{fig:visual_1} and Fig.~\ref{fig:visual_2}), thereby facilitating the training pipeline in also satisfying \ul{Attribute 2}. Moreover, it is appealingly efficient, incurring only minimal additional time cost, as nearly all of its required operations can be executed in parallel with existing steps (i.e., the meta-training and meta-testing steps). At this point, we have derived both the meta-testing feedback and the meta-flatness feedback. In the final step of our scheme (\textbf{actual optimization}), we jointly leverage these two feedbacks to adjust the simulated training step of the generator and actually optimize it. By integrating both feedbacks, this step guides the actual optimization of the generator to simultaneously satisfy \ul{Attribute 1} and \ul{Attribute 2}. In doing so, it facilitates the generator in learning to generate broadly unexploitable examples effectively. Below, we formally describe the above-outlined four steps.

\noindent\textbf{Meta-training.} Denote $f_s^{m\_{tr}}$ a surrogate model that is randomly chosen from the meta-training set $D^{m\_{tr}}$. In this step, we train the generator $f_g$ for one step leveraging $f_s^{m\_{tr}}$. In specific, based on Eq.~\ref{eq:basic_3}, we first calculate the loss $L^{m\_{tr}}$ w.r.t. this training step as:
\begin{equation}\label{eq:meta_1}
\begin{aligned}
L^{m\_{tr}}(f_g) = L^{m\_{tr}}_s\Big(f^{m\_{tr}}_s\big(\mathit{tanh}(f_g(x)) \times \epsilon + x\big)\Big), 
\end{aligned}
\end{equation}
where $L^{m\_{tr}}_s$ is the loss function associated with $f^{m\_{tr}}_s$. Based on $L^{m\_{tr}}(f_g)$, we can then obtain the corresponding meta-training gradient $\nabla_{f_g} L^{m\_tr}(f_g)$ and train $f_g$ via gradient descent as:
\begin{equation}\label{eq:meta_2}
\begin{aligned}
f_g^{desc} = f_g - \alpha \nabla_{f_g} L^{m\_tr}(f_g),
\end{aligned}
\end{equation}
where $\alpha$ denotes the learning rate for meta-training. It is worth noting that what we perform here is just to train $f_g$ simulatedly. In other words, the updated model $f_g^{desc}$ obtained in this step serves merely as a temporary intermediary (that will be used in the next meta-testing step), and we will actually optimize the generator $f_g$ in the actual-optimization step.

\noindent\textbf{Meta-testing.} After meta-training the generator $f_g$ and obtaining $f_g^{desc}$, we proceed to evaluate the cross-task generalization performance of $f_g^{desc}$ on the meta-testing set $D^{m\_{te}}$. In specific, denote $f_s^{m\_{te}}$ a surrogate model randomly selected from $D^{m\_{te}}$. We perform such cross-task evaluation as:
\begin{equation}\label{eq:meta_3}
\begin{aligned}
L^{m\_{te}}(f_g^{desc}) =L^{m\_{te}}_s\Big(f^{m\_{te}}_s\big(\mathit{tanh}(f_g^{desc}(x)) \times \epsilon + x\big)\Big),
\end{aligned}
\end{equation}
where $L^{m\_{te}}_s$ represents the loss function associated with $f^{m\_{te}}_s$. Based on $L^{m\_{te}}(f_g^{desc})$, we can then formulate the \textbf{meta-testing feedback} $g_{meta-test}$ as:
\begin{equation}\label{eq:meta_4}
\begin{aligned}
g_{meta-test} & = \nabla_{f_g} L^{m\_{te}}(f_g^{desc}) \\
& = \nabla_{f_g} L^{m\_{te}}(f_g - \alpha \nabla_{f_g} L^{m\_tr}(f_g)).
\end{aligned}
\end{equation}

We now explain why $g_{meta-test}$ formulated in Eq.~\ref{eq:meta_4} can effectively guide the generator to learn more cross-task unexploitable knowledge. Specifically, recall that during meta-testing and correspondingly formulating $g_{meta-test}$, the evaluation is performed on $f_g^{desc}$, which is obtained via simulated training $f_g$ leveraging $f^{m\_{tr}}_s$. Consequently, $f_g^{desc}$ can possibly learn task-specific knowledge that only works when unexploitable examples are generated for $f^{m\_{tr}}_s$. However, to generalize well to the surrogate model $f^{m\_{te}}_s$ where task-specific knowledge tailored to $f^{m\_{tr}}_s$ is no longer applicable, the generator needs to learn to avoid acquiring task-specific knowledge, and instead focus on capturing more cross-task unexploitable knowledge. 

The above means that, $g_{meta-test}$, as the second-order gradient of $f_g$ (i.e., $\nabla_{f_g} L^{m\_{te}}(f_g - \alpha \nabla_{f_g} L^{m\_tr}(f_g))$), calculated from the evaluation of $f_g^{desc}$ on $f^{m\_{te}}_s$, can be interpreted as a form of \textit{cross-task generalization} feedback. In other words, it can provides informative feedback to the generator's training process ($f_g \rightarrow f_g^{desc}$), about how to learn more cross-task unexploitable knowledge. This explains why incorporating $g_{meta-test}$ formulated in the above way into the training pipeline of $f_g$ can effectively facilitate the pipeline in satisfying \ul{Attribute 1}. Notably, besides the above intuitive explanation, we also provide a theoretical analysis of the effectiveness of $g_{meta-test}$ in Supplementary.

\noindent\textbf{Meta-flatness-assessment.} Above, we introduced the formulation of the meta-testing feedback $g_{meta-test}$, which can be used to effectively guide the training process of $f_g$ to satisfy \ul{Attribute 1}. Here, we aim to steer the optimization of $f_g$ to further satisfy \ul{Attribute 2}—that is, to converge $f_g$ further towards flat minima in the loss landscape, thereby enhancing its ability to robustly preserve its learned (cross-task unexploitable) knowledge even under distribution shifts. To achieve this, a naive approach might involve directly using the gradient norm of the loss function or the maximal eigenvalue of the Hessian matrix to additionally regularize the generator during its optimization \cite{kaur2023maximum,zhang2023gradient}. Such an approach, however, may be suboptimal in our case. This is because, in our optimization process, which involves numerous parameters spanning multiple models (both $f_g$ and $\{f^t_s\}^T_{t=1}$), the computation required behind the above naive approach may be extremely expensive or even infeasible given current hardware constraints \cite{sankar2021deeper,an2021can,zhao2022penalizing}. In light of this, rather than directly adopting existing techniques, we aim to satisfy \ul{Attribute 2} in our training pipeline by carefully leveraging its inherent characteristics. 
Our goal is to encourage the training pipeline's compliance with \ul{Attribute 2} while incurring only minimal additional time overhead.

To achieve this, we begin with the following key observation: \textit{the meta-training gradient term $\nabla_{f_g} L^{m\_tr}(f_g)$, already computed during the meta-training step, encodes valuable information about the local flatness of the loss landscape around $f_g$}. Specifically, it suggests the direction along which the loss $L^{m\_tr}(f_g)$ increases most steeply from $f_g$. Motivated by this, we propose taking a single ascent step along this direction, reaching at a new point denoted as $f_g^{asc}$. The difference (gap) in loss values between $f_g$ and the intermediary point $f_g^{asc}$ then serves as an effective proxy for assessing the flatness of the loss landscape surrounding $f_g$ \cite{foret2021sharpnessaware}, a smaller gap implies a flatter surrounding region. As a result, regularizing this gap offers a principled way of guiding $f_g$ toward points with flat surrounding regions in the landscape (i.e. flat minima). Formally, we derive $f_g^{asc}$ and correspondingly define this gap $L^{m\_tr}_{gap}(f_g^{asc}, f_g)$ as:
\begin{equation}\label{eq:sharp1}
\begin{aligned}
f_g^{asc} = f_g +\eta \nabla_{f_g} L^{m\_{tr}}(f_g),
\end{aligned}
\end{equation}
\begin{equation}\label{eq:sharp2}
\begin{aligned}
L^{m\_tr}_{gap}(f_g^{asc}, f_g) =  L^{m\_tr}(f_g^{asc})-L^{m\_tr}(f_g),
\end{aligned}
\end{equation}
where $\eta$ denotes the gradient ascent step size used for meta-flatness assessment. Importantly, the computations in Eq.~\ref{eq:sharp1} and Eq.~\ref{eq:sharp2} can be performed in parallel with those in Eq.~\ref{eq:meta_2} and Eq.~\ref{eq:meta_3}, respectively. This not only makes $L^{m\_tr}_{gap}(f_g^{asc}, f_g)$ an effective flatness indicator, but also leads its computation to require only minimal additional time cost during the optimization process of $f_g$.

Above, we derive the gap term $L^{m\_tr}_{gap}(f_g^{asc}, f_g)$, whose gradient $\nabla_{f_g} L^{m\_tr}_{gap}(f_g^{asc}, f_g)$ can already serve as a flatness feedback, converging $f_g$ toward flat minima in the loss landscape. However, relying solely on $\nabla_{f_g} L^{m\_tr}_{gap}(f_g^{asc}, f_g)$ to regularize $f_g$ may be suboptimal, as it reflects only the loss landscape of the current meta-training task while ignoring the remaining tasks. This narrow focus can lead to a biased assessment of flatness, and hinder convergence toward the common flat minima across different tasks \cite{zhang2024domaininspired}. To tackle this and better satisfy \ul{Attribute 2}, we propose further augmenting $\nabla_{f_g} L^{m\_tr}_{gap}(f_g^{asc}, f_g)$ with flatness feedback signals derived from the other tasks. 

To achieve this, a naive strategy would be to treat each task in turn as the meta-training task and compute its corresponding flatness feedback signal (i.e., $\nabla_{f_g} L^{m\_tr}_{gap}(f_g^{asc}, f_g)$) one by one. However, this approach can be computationally demanding and thus undesirable, as it requires performing the meta-training step $T$ times per iteration—once for each seen task. Fortunately, our optimization process possesses an appealing characteristic: \textit{it naturally cycles through different tasks as meta-training tasks across iterations and epochs}. We also draw inspiration from \cite{schmidt2017minimizing,defazio2014saga,chai2024towards}, which show that averaging gradients of the same objective across past iterations yields an efficient approximation when recomputing that gradient in the current iteration. 
Leveraging these insights, we propose to reuse flatness feedback signals computed in the past iterations when other tasks were used as meta-training tasks to augment the current feedback. This leads to the formulation of our final \textbf{meta-flatness feedback}, denoted as $g_{meta-flat}$, which efficiently aggregates flatness information across tasks. Formally, let $S^{rest}$ denote the set of surrogate models for all seen tasks except the current meta-training task (i.e., $S^{rest} = \{f^t_s\}^T_{t=1} \setminus f^{m\_tr}_s$). We define $g_{meta-flat}$ as:
\begin{equation}\label{eq:sharp5}
\begin{aligned}
g_{meta-flat}=& \nabla_{f_g} L^{m\_tr}_{gap}(f_g^{asc}, f_g) \\
& + \lambda \sum_{f^t_s \in S^{rest}} \nabla_{f_g} L^{t,p\_avg}_{gap}(f_g^{asc}, f_g),
\end{aligned}
\end{equation}
where, with slight abuse of notation, $\nabla_{f_g} L^{t,p\_avg}_{gap}(f_g^{asc}, f_g)$ represents the average of the flatness feedback signals computed from past iterations in which $f^t_s$ was used as the surrogate model during meta-training (see Supplementary for details), and $\lambda$ is a weighting hyperparameter. 
By ultimately formulating $g_{meta-flat}$ in this way, via incorporating it into the actual optimization of $f_g$, we enable guiding $f_g$ toward the common flat minima across tasks in both a effective and efficient manner, thereby more properly satisfying \underline{Attribute 2}.

\noindent\textbf{Actual-optimization.} At this point, following the steps outlined above, we have derived the gradient term $\nabla_{f_g}L^{m\_tr}(f_g)$ from the simulated-training process of the generator $f_g$ (Eq.~\ref{eq:meta_2}). In addition, we have obtained two feedback signals: the meta-testing feedback $g_{\text{meta-test}}$, derived from the meta-testing step and capable of encouraging satisfaction of \ul{Attribute 1} when incorporated into the simulated-training process of $f_g$; and the meta-flatness feedback $g_{\text{meta-flat}}$, derived from the meta-flatness-assessment step and capable of further encouraging satisfaction of \ul{Attribute 2}. Our final objective is to jointly leverage these feedback signals to refine the simulated-training process of the generator $f_g$ and perform the actual optimization of $f_g$. Formally, this leads us to the following update rule:
\begin{equation}\label{eq:meta_5}
\begin{aligned}
f_g \leftarrow & f_g - \beta\big( \nabla_{f_g}L^{m\_{tr}}(f_g) + g_{meta-test} + g_{meta-flat}\big),
\end{aligned}
\end{equation}
where $\beta$ is the learning rate used for the actual optimization. By updating $f_g$ in this way, we can finally guide the actual optimization of the generator to simultaneously satisfy \ul{Attribute 1} and \ul{Attribute 2}, ultimately promoting the practical and effective generation of broadly unexploitable examples. 

\subsection{Overall Training and Testing}
\label{sec:overall}

During training, our framework alternatively optimizes the unexploitable example generator $f_g$ and the surrogate models $\{f^t_s\}^T_{t=1}$. In specific, in each epoch where $f_g$ is to be optimized, at start, we first split $\{f^t_s\}^T_{t=1}$ to construct the  meta-training and meta-testing sets (as introduced in Sec.~\ref{sec:meta}). Then, in each iteration, we optimize $f_g$ using the actual-optimization rule in Eq.~\ref{eq:meta_5}. In epochs optimizing $\{f^t_s\}^T_{t=1}$, we follow the approach detailed in Sec.~\ref{sec:base}. During testing, we use the trained generator to produce unexploitable examples in the same way as \cite{sun2024unseg}. 

\section{Experiments}
\label{sec:experiments}

Previous unexploitable example generation work UnSeg \cite{sun2024unseg}  proposed a multi-task evaluation protocol (\textbf{Protocol A}). In this protocol, the evaluation is made on a limited range of computer vision tasks relevant to the Segment Anything Model \cite{kirillov2023segment}, focusing primarily on segmentation tasks. To assess the efficacy of our cross-task unexploitable example generation framework MCT-UEG, we first conduct experiments following this protocol. Additionally, to enable a more comprehensive evaluation across a broader variety of computer vision tasks, we also propose a new evaluation protocol (\textbf{Protocol B}), which evaluates on the Taskonomy dataset \cite{zamir2018taskonomy}. This dataset provides annotations for a wide variety of tasks, allowing us to evaluate our framework further across more diverse (unseen) tasks that can appear in practice. We conduct our experiments on RTX 4090 GPUs. 

\noindent\textbf{Evaluation Pipeline.} In both protocols, after training the unexploitable example generator, we evaluate its efficacy for a given computer vision task and dataset by following the pipeline outlined in \cite{sun2024unseg}, which comprises three steps: 1) First, we use the trained generator to transform each image in the training set of the evaluation dataset into its unexploitable version. 2) These transformed examples are then used to train a target model for the specified computer vision task. 3) Finally, the performance of the trained target model is evaluated on the testing set of the evaluation dataset. Notably, \ul{worse performance of the trained target model suggests that it has learned less useful information from the unexploitable examples, thereby indicating higher efficacy of the unexploitable example generator.}

\begin{table}[t]
\caption{Performance comparison under Protocol A.}
\centering
\resizebox{\columnwidth}{!}
{
\setlength{\tabcolsep}{2pt}
\begin{tabular}{l|ccccccccccc} \hline
\multirow{2}{*}{\textbf{Method}} &  \multicolumn{3}{c}{\textbf{Panoptic Segmentation}} & \multicolumn{4}{c}{\textbf{Instance Segmentation}} &  \multicolumn{4}{c}{\textbf{Object Detection}} \\ 
\cmidrule(lr){2-4} \cmidrule(lr){5-8} \cmidrule(lr){9-12} 
& PQ & $AP^{Th}_{pan}$ & $mIoU_{pan}$ & $AP$ & $AP^S$ & $AP^M$ & $AP^L$ & $AP$ & $AP^S$ & $AP^M$ & $AP^L$ \\ \hline\hline
Raw data & 51.9 & 41.7 & 61.7 & 43.7 & 23.4 & 47.2 & 64.8 & 48.7 & 31.1 & 51.9 & 62.9 \\
\hline
Random noise & 47.1 & 38.1 & 57.7 & 39.5 & 19.7 & 42.1 & 60.7 & 47.4 & 30.9 & 50.6 & 60.6 \\
SynPer \cite{yu2022availability} & 11.3 & 9.5 & 11.0 & 10.8 & 13.4 & 15.2 & 5.0 & 38.9 & 24.4 & 42.9 & 50.9 \\
CUDA \cite{sadasivan2023cuda} & 6.7 & 4.7 & 11.2 & 9.7 & 3.7 & 10.9 & 18.8 & 46.8 & 29.3 & 50.6 & 61.2 \\
UnSeg \cite{sun2024unseg} & 4.2 & 3.2 & 5.2 & 4.0 & 5.8 & 3.7 & 1.7 & 6.1 & 9.3 & 5.7 & 2.4 \\
\hline
Ours & \textbf{3.1} & \textbf{2.8} & \textbf{4.4} & \textbf{3.3} & \textbf{3.4} & \textbf{3.1} & \textbf{1.2} & \textbf{3.9} & \textbf{4.9} & \textbf{4.7} & \textbf{2.2} \\
\hline
\end{tabular}}
\label{Tab:unseg}
\end{table}

\subsection{Protocol A}
\label{sec:unseg_protocol}

\noindent\textbf{Experimental Setting.} In this protocol, following \cite{sun2024unseg}, we first train our unexploitable example generator on the HQSeg-44K \cite{ke2024segment} dataset. During training, similar to \cite{sun2024unseg}, we set $T=6$ and expose our generator to six tasks, including point-based, box-based, mask-based, point-and-box-based, point-and-mask-based, and box-and-mask-based interactive segmentation. After training, following \cite{sun2024unseg}, we evaluate the trained generator using the above-introduced pipeline, on the COCO dataset \cite{lin2014microsoft} and across three different tasks: panoptic segmentation, instance segmentation, and object detection. Here, we adopt the same train/test split of the COCO dataset as in \cite{sun2024unseg}. Moreover, consistent with \cite{sun2024unseg}, during the evaluation on panoptic segmentation, we use Mask2Former \cite{cheng2022masked} as the target model and report the following metrics: Panoptic Quality (PQ) \cite{kirillov2019panoptic}, $AP^{Th}_{pan}$, and $mIoU_{pan}$. For instance segmentation evaluation, we also use Mask2Former and we report metrics including $AP$, $AP^S$, $AP^M$, and $AP^L$. For object detection evaluation, we use DINO \cite{zhang2022dino} as the target model and report the same four metrics as for instance segmentation. Both Mask2Former \cite{cheng2022masked} and DINO \cite{zhang2022dino} are employed with a ResNet50 backbone, maintaining consistency with \cite{sun2024unseg}. Additionally, we set the maximum allowable magnitude $\epsilon$ in the same manner as \cite{sun2024unseg} during both training and evaluation.

\noindent\textbf{Implementation Details.} Our framework uses the same model architecture as \cite{sun2024unseg} for both the unexploitable example generator and the surrogate models. Consistent with \cite{sun2024unseg}, we adopt the following alternating update strategy during training: the generator is updated for three consecutive epochs, followed by one epoch of updates for the surrogate models.
This alternating process is repeated until convergence. We set the hyperparameters $\alpha$ to 1e-4, $\beta$ to 2e-5, $\eta$ to 5e-4, and $\lambda$ to 0.2.

\noindent\textbf{Results.} 
In Tab.~\ref{Tab:unseg}, we compare our framework with existing unexploitable example generation methods that are free of per-dataset optimization and require no external information beyond the testing images during evaluation (e.g., text descriptions corresponding to each testing image). We also compare our framework with Random noise as a baseline that introduces a random noise of magnitude $\epsilon$ to each image to generate its unexploitable version. As shown, our framework consistently outperforms both existing methods and the Random noise baseline across all tasks and metrics, showing its cross-task effectiveness.

\begin{table*}[t]
\caption{Performance comparison under Protocol B, which contains \textbf{more diverse (unseen) tasks}.}
\centering
\resizebox{\textwidth}{!}
{
\setlength{\tabcolsep}{2pt}
\begin{tabular}{l|cccc} \hline
\textbf{Method} & \makecell{\textbf{Semantic Segmentation~(mIoU$\downarrow$)}}  & \makecell{\textbf{Scene Recognition~(accuracy$\downarrow$)}} & \makecell{\textbf{Keypoint Detection~(MAE$\uparrow$)}} & \makecell{\textbf{Image Inpainting~(FID$\uparrow$)}} \\
\hline\hline
Raw data & 41.1 & 71.0 & 3.1 & 59.3 \\
\hline
Random noise & 39.0 & 69.4 & 3.2 & 60.1 \\
SynPer \cite{yu2022availability} & 39.5 & 69.8 & 3.2 & 60.4 \\
CUDA \cite{sadasivan2023cuda} & 41.0 & 68.1 & 3.3 & 63.9 \\
\hline
Unseg \cite{sun2024unseg} (Experiment B.1) & 36.6 & 68.9 & 4.9 & 61.5 \\
Ours (Experiment B.1) & \textbf{21.5} & \textbf{45.2} & \textbf{16.5} & \textbf{80.4} \\
\hline
UnSeg \cite{sun2024unseg} (Experiment B.2) & 9.2 & 61.9 & 5.5 & 62.7 \\
Ours (Experiment B.2) & \textbf{7.5} & \textbf{33.9} & \textbf{22.5} & \textbf{133.2} \\
\hline
\end{tabular}}
\label{Tab:taskonomy}
\end{table*}

\subsection{Protocol B}

\noindent\textbf{Experimental Setting.} In this protocol, we evaluate our framework more comprehensively across a broader variety of computer vision tasks. In specific, we conduct two groups of experiments. 
1) In \textbf{Experiment B.1}, we do not train a new generator. Instead, we further assess the generator trained in Protocol A (i.e., trained on the HQSeg-44K dataset \cite{ke2024segment} and on six interactive segmentation tasks) across a broader array of previously unseen computer vision tasks. This evaluation is carried out on the Taskonomy dataset \cite{zamir2018taskonomy}, covering four diverse tasks: semantic segmentation, scene recognition, keypoint detection, and image inpainting. The target models for these tasks are constructed following the setup in \cite{zamir2018taskonomy}. 2) In \textbf{Experiment B.2}, we train a new generator to further explore the scenario where the generator is exposed to a mix of common computer vision tasks rather than only segmentation-type tasks during training. Specifically, we train the generator on the train2017 subset of the COCO dataset \cite{lin2014microsoft}, allowing it to encounter tasks including semantic segmentation, object detection, and human pose estimation during training. After training, we evaluate the generator using the same procedure as in Experiment B.1. In both Experiment B.1 and Experiment B.2, following \cite{huang2021unlearnable}, we set the maximum allowable noise magnitude $\epsilon$ to $8/255$ during evaluation. We use the Taskonomy-tiny train/test split of the Taskonomy dataset, which contains around 25k training images and 5k testing images. Besides, we report the following evaluation metrics: mIoU for semantic segmentation, accuracy for scene recognition, MAE for keypoint detection, and Fréchet Inception Distance (FID) \cite{heusel2017gans} for image inpainting. In this context, lower mIoU, lower accuracy, higher MAE, and higher FID indicate worse performance of the target models, which in turn reflects better efficacy of the unexploitable
example generation method, as explained in the above evaluation pipeline part.

\noindent\textbf{Implementation Details.} In Experiment B.2, we adopt the widely used YOLO architecture, specifically its recent YOLO 11 version \cite{Jocher_Ultralytics_YOLO_2023}, for the surrogate models. For a fair comparison, this same architecture is employed by both UnSeg \cite{sun2024unseg} and our framework in Experiment B.2. All other implementation details follow those in Protocol A.

\noindent\textbf{Results.} In Tab.~\ref{Tab:taskonomy}, we present the results of both Experiment B.1 and Experiment B.2 under protocol B. As shown, on both groups of experiments, our framework consistently outperforms all previous methods significantly, across diverse types of evaluation tasks with varying characteristics. This further underscores the cross-task effectiveness of our framework.

\subsection{Ablation Studies}

In this section, we conduct extensive ablation studies in Experiment B.1 under Protocol B. \textbf{More ablation studies are in Supplementary.}

\noindent\textbf{Impact of main designs of our framework.} In our framework, we introduce a \textit{flat-minima-oriented meta training and testing scheme} to optimize the generator. To systematically validate the efficacy of this scheme, we test two variants. In the first variant (\textbf{w/o meta-testing feedback}), we remove the meta-testing feedback from Eq.~\ref{eq:meta_5} and only adjust the simulated-training process of the generator leveraging the meta-flatness feedback. In the second variant (\textbf{w/o meta-flatness feedback}), we remove the meta-flatness feedback instead. As shown in Tab.~\ref{Tab:ablation_1}, our MCT-UEG framework obviously outperforms both variants, showing the efficacy of both main designs we made in our framework. 

\begin{table}[h]
\caption{Evaluation on main designs of our framework.} 
\centering
\resizebox{\columnwidth}{!}
{
\setlength{\tabcolsep}{2pt}
\begin{tabular}{l|cccc} \hline
\textbf{Method} & \makecell{\textbf{Semantic} \\ \textbf{Segmentation} \\ \textbf{(mIoU$\downarrow$)}} & \makecell{\textbf{Scene} \\ \textbf{Recognition}  \\ \textbf{(accuracy$\downarrow$)}} & \makecell{\textbf{Keypoint} \\ \textbf{Detection} \\ \textbf{(MAE$\uparrow$)}} & \makecell{\textbf{Image} \\ \textbf{Inpainting} \\ \textbf{(FID$\uparrow$)}} \\
\hline
Baseline (UnSeg) & 36.6 & 68.9 & 4.9 & 61.5 \\
\hline
w/o meta-testing feedback & 32.5 & 57.0 & 10.3 & 69.7 \\
w/o meta-flatness feedback & 29.5 & 54.0 & 12.5 & 74.8 \\
MCT-UEG  & 21.5 & 45.2 & 16.5 & 80.4 \\
\hline
\end{tabular}}
\label{Tab:ablation_1}
\end{table}

\noindent\textbf{Time analysis.} In Tab.~\ref{Tab:ablation_5.5}, we evaluate the training time of our framework and compare it to the training time of the baseline method UnSeg \cite{sun2024unseg}, on four RTX 4090 GPUs in terms of hours. As shown, though our method achieves much better performance, it brings only relatively little increase ($9.9\%$) of the training time. 
Meanwhile, the testing time of our framework remains identical to that of the baseline method UnSeg, since our framework shares the same testing pipeline as UnSeg (as mentioned in Sec.~\ref{sec:overall}).

\begin{table}[H]
\caption{Comparison of the training time.}
\centering
\resizebox{0.9\columnwidth}{!}
{
\setlength{\tabcolsep}{2pt}
\begin{tabular}{l|ccccc} \hline
\textbf{Method} &  \textbf{Semantic Segmentation(mIoU$\downarrow$)} & \textbf{Training time}\\
\hline
Baseline(UnSeg) & 36.6 & 29.2h\\
\hline
MCT-UEG & 21.5 & 32.1h\\
\hline
\end{tabular}}
\label{Tab:ablation_5.5}
\end{table}

\noindent\textbf{Impact of the meta-testing feedback $g_{meta-test}$}. In our framework, to facilitate the unexploitable example generator in satisfying \ul{Attribute 1}, we propose a second-order-gradient-based meta-testing feedback $g_{meta-test}$, where $g_{meta-test} = \nabla_{f_g}L^{m\_{te}}\big(f_g - \alpha \nabla_{f_g} L^{m\_tr}(f_g)\big)$. To evaluate the efficacy of this feedback, we test two variants. In the first variant (\textbf{w/o meta-testing feedback}), we remove the meta-testing feedback $g_{meta-test}$ from the optimization process. In the second variant (\textbf{w/o second-order gradient}), we use $\nabla_{f_g} \big( L^{m\_{te}}(f_g) \big)$ instead of $\nabla_{f_g}L^{m\_{te}}\big(f_g - \alpha \nabla_{f_g} L^{m\_tr}(f_g)\big)$ as the meta-testing feedback in the optimization process. As shown in Tab.~\ref{Tab:ablation_1_supp}, compared to our MCT-UEG framework, the performance of both variants drops significantly. This shows the efficacy of our meta-testing feedback design. 

\begin{table}[h]
\caption{Evaluation on the meta-testing feedback $g_{meta-test}$.}
\centering
\resizebox{\columnwidth}{!}
{
\setlength{\tabcolsep}{2pt}
\begin{tabular}{l|cccc} \hline
\textbf{Method} & \makecell{\textbf{Semantic} \\ \textbf{Segmentation} \\ \textbf{(mIoU$\downarrow$)}} & \makecell{\textbf{Scene} \\ \textbf{Recognition}  \\ \textbf{(accuracy$\downarrow$)}} & \makecell{\textbf{Keypoint} \\ \textbf{Detection} \\ \textbf{(MAE$\uparrow$)}} & \makecell{\textbf{Image} \\ \textbf{Inpainting} \\ \textbf{(FID$\uparrow$)}} \\
\hline
w/o meta-testing feedback &  32.5 & 57.0 & 10.3 & 69.7 \\
w/o second order gradient &  31.6 & 56.1 & 11.3 & 70.1 \\
\hline
MCT-UEG & 21.5 & 45.2 & 16.5 & 80.4 \\
\hline
\end{tabular}}
\label{Tab:ablation_1_supp}
\end{table}

\noindent\textbf{Impact of the meta-flatness feedback $g_{meta-flat}$}. In our framework, to facilitate the unexploitable example generator in also satisfying \ul{Attribute 2}, we further design a meta-flatness feedback $g_{meta-flat}$. In specific, as shown in Eq.~\ref{eq:sharp5}, $g_{meta-flat}$ involves the flatness feedback signal $\nabla_{f_g} L^{m\_tr}_{gap}(f_g^{asc}, f_g)$ computed from the meta-training surrogate model $f_s^{m\_tr}$ in the current iteration, while further augments $\nabla_{f_g} L^{m\_tr}_{gap}(f_g^{asc}, f_g)$ with the historical average flatness feedback signals $\sum_{f^t_s \in S^{rest}} \nabla_{f_g} L^{t,p\_avg}_{gap}(f_g^{asc}, f_g)$. To validate our design of the meta-flatness feedback $g_{meta-flat}$, we test two variants. In the first variant (\textbf{w/o meta-flatness feedback}), we remove the whole meta-flatness feedback from the optimization process. In the second variant (\textbf{w/o historical average flatness feedback signals}), we still utilize the meta-flatness feedback in the optimization process but remove its $\sum_{f^t_s \in S^{rest}} \nabla_{f_g} L^{t,p\_avg}_{gap}(f_g^{asc}, f_g)$ component. As shown in Tab~\ref{Tab:ablation_4}, our MCT-UEG framework outperforms both variants. Meanwhile, compared to both variants, our framework does not increase much training time. This shows both the effectiveness and the efficiency of our designed meta-flatness feedback $g_{meta-flat}$.

\begin{table}[h]
\caption{Evaluation on the meta-flatness feedback $g_{meta-flat}$.}
\centering
\resizebox{\columnwidth}{!}
{
\setlength{\tabcolsep}{2pt}
\begin{tabular}{l|ccccc} 
\hline
\textbf{Method} & \makecell{\textbf{Semantic} \\ \textbf{Segmentation} \\ \textbf{(mIoU$\downarrow$)}} & \makecell{\textbf{Scene} \\ \textbf{Recognition}  \\ \textbf{(accuracy$\downarrow$)}} & \makecell{\textbf{Keypoint} \\ \textbf{Detection} \\ \textbf{(MAE$\uparrow$)}} & \makecell{\textbf{Image} \\ \textbf{Inpainting} \\ \textbf{(FID$\uparrow$)}} & \makecell{\textbf{Training} \\ \textbf{time}} \\
\hline 
w/o meta-flatness feedback  & 29.5 & 54.0 & 12.5 & 74.8 & 30.0h \\
w/o historical average flatness feedback signals & 24.7 & 48.2 & 15.2 & 78.2 & 31.6h  \\
\hline
MCT-UEG & 21.5 & 45.2 & 16.5 & 80.4 & 32.1h \\
\hline
\end{tabular}}
\label{Tab:ablation_4}
\end{table}

\noindent\textbf{Analysis on the feasibility of existing naive flatness regularizers.} In our framework, we do not directly adopt existing naive flatness regularizers, like the gradient norm of the loss function or the maximal eigenvalue of the Hessian matrix, to regularize the generator. Instead, we design a meta-flatness feedback and use it to converge the unexploitable example generator to flat minima in the loss landscape. Here, to further analyze on this choice, we test two variants. In the first variant (\textbf{gradient norm}), instead of our meta-flatness feedback, we directly use the gradient norm of the loss function to regularize the generator. In the second variant (\textbf{maximum eigenvalue}), rather than our meta-flatness feedback, we directly adopt the maximum eigenvalue of the Hessian matrix to regularize the generator. As shown in Tab~\ref{Tab:ablation_3}, both these two variants incur out-of-memory (OOM) errors, showing the infeasibility of these variants. We perform this analysis on an RTX 4090 GPU with 24 GB of GPU memory, using a minimal batch size of 1. 

\begin{table}[h]
\caption{Evaluation on the feasibility of existing naive flatness regularizers.}
\centering
\resizebox{0.7\columnwidth}{!}
{
\setlength{\tabcolsep}{2pt}
\begin{tabular}{l|cc} \hline
\textbf{Method} & \textbf{Approximate GPU Memory} \\
\hline
Gradient norm & OOM\\
Maximum eigenvalue & OOM\\
\hline
MCT-UEG & 8.86G\\
\hline
\end{tabular}}
\label{Tab:ablation_3}
\end{table}

\noindent\textbf{Impact of seeking help from past iterations when performing flatness regularization across tasks.} In our framework, when formulating $g_{meta-flat}$, during the augmentation of $\nabla_{f_g} L^{m\_tr}_{gap}(f_g^{asc}, f_g)$ with flatness feedback signals derived from the other seen tasks (besides the meta-training task in the current iteration), we propose seeking help from past iterations in the optimization process, instead of naively re-computing all task's corresponding flatness feedback signals one by one in every iteration. To validate this design choice, we test a variant (\textbf{w/o seeking help from past iterations}) that directly adopts the above-mentioned naive strategy. As shown in Tab.~\ref{Tab:ablation_5}, our framework (\textbf{with seeking help from past iterations}) requires significantly less training time than this variant while achieving comparable performance. This demonstrates the benefit of seeking help from past iterations when formulating $g_{meta-flat}$.

\begin{table}[H]
\caption{Evaluation on seeking help from past iterations when performing flatness regularization across tasks.}
\centering
\resizebox{\columnwidth}{!}
{
\setlength{\tabcolsep}{2pt}
\begin{tabular}{l|ccccc} \hline

\textbf{Method} & \makecell{\textbf{Training time}} & \makecell{\textbf{Semantic Segmentation} \\ \textbf{(mIoU$\downarrow$)}} \\
\hline
w/o seeking help from past iterations  & 97.7h & 21.6 \\
with seeking help from past iterations & 32.1h & 21.5 \\
\hline
\end{tabular}}
\label{Tab:ablation_5}
\end{table}

\definecolor{myblue}{RGB}{19,111,176}

\begin{figure}[t]
\centering
\includegraphics[width=1\columnwidth]{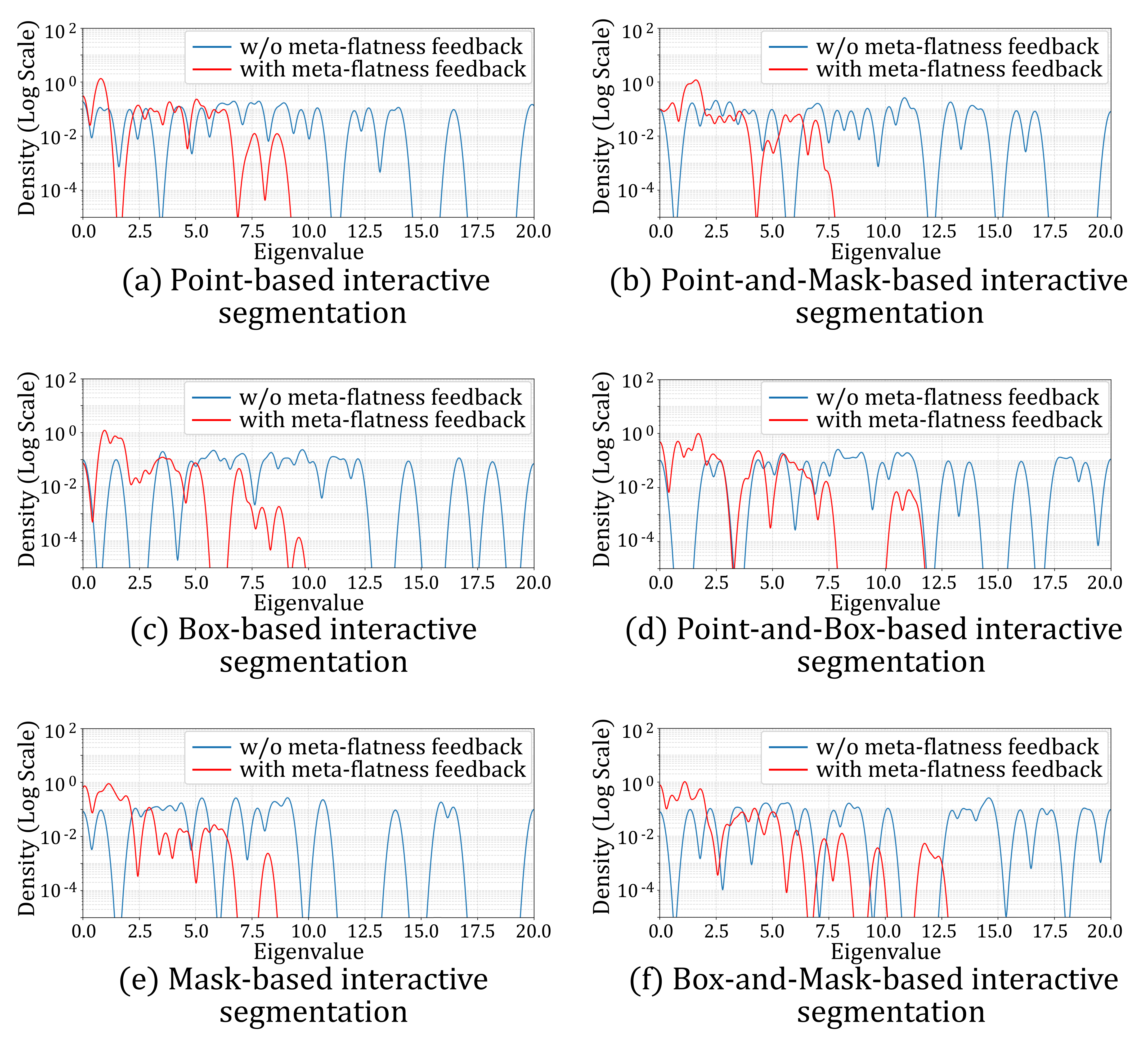}
\caption{Visualization on the Hessian spectra calculated over seen tasks, on the HQSeg-44K dataset \cite{ke2024segment}. Notably, a Hessian spectrum whose peaks are more concentrated on the left-hand side (i.e., more concentrated on small eigenvalues) represents flatter slopes around in the loss landscape.} 
\label{fig:visual_1}
\end{figure}

\noindent\textbf{Visualization of minima flatness.} In our framework, to guide the generator to robustly retain its learned cross-task unexploitable knowledge under distribution shifts, we propose a meta-flatness-assessment step, in which we design a meta-flatness feedback $g_{meta-flat}$ to converge the unexploitable example generator toward flat minima in the loss landscape. To show that this feedback indeed helps converge the generator toward flat minima with flat slopes around, inspired by \cite{foret2021sharpnessaware} that a Hessian spectrum with peaks more concentrated on its left-hand side (i.e., more concentrated on small eigenvalues) indicates flatter slopes around in the loss landscape, following \cite{foret2021sharpnessaware} and using the Lanczos algorithm, we here estimate the Hessian spectra of the minima. Specifically, each sub-figure in Fig.~\ref{fig:visual_1} displays two Hessian spectra computed from the loss landscape of one specific seen task: the \textcolor{red}{red} curve shows the spectrum obtained from our framework (\textbf{with meta-flatness feedback}), and the \textcolor{myblue}{blue} curve shows the spectrum obtained from the variant \textbf{w/o meta-flatness feedback} defined in Tab.~\ref{Tab:ablation_4}. All spectra in Fig.~\ref{fig:visual_1} are computed on the HQSeg-44K dataset \cite{ke2024segment} after the last optimization epoch. As shown, with our meta-flatness feedback, the Hessian spectra consistently exhibit peaks more concentrated on the left-hand side across different seen tasks. This indicates that our meta-flatness feedback $g_{meta-flat}$ indeed helps the unexploitable example generator converge toward common flat minima across these tasks.

\begin{figure}[t]
\centering
\includegraphics[width=1\columnwidth]{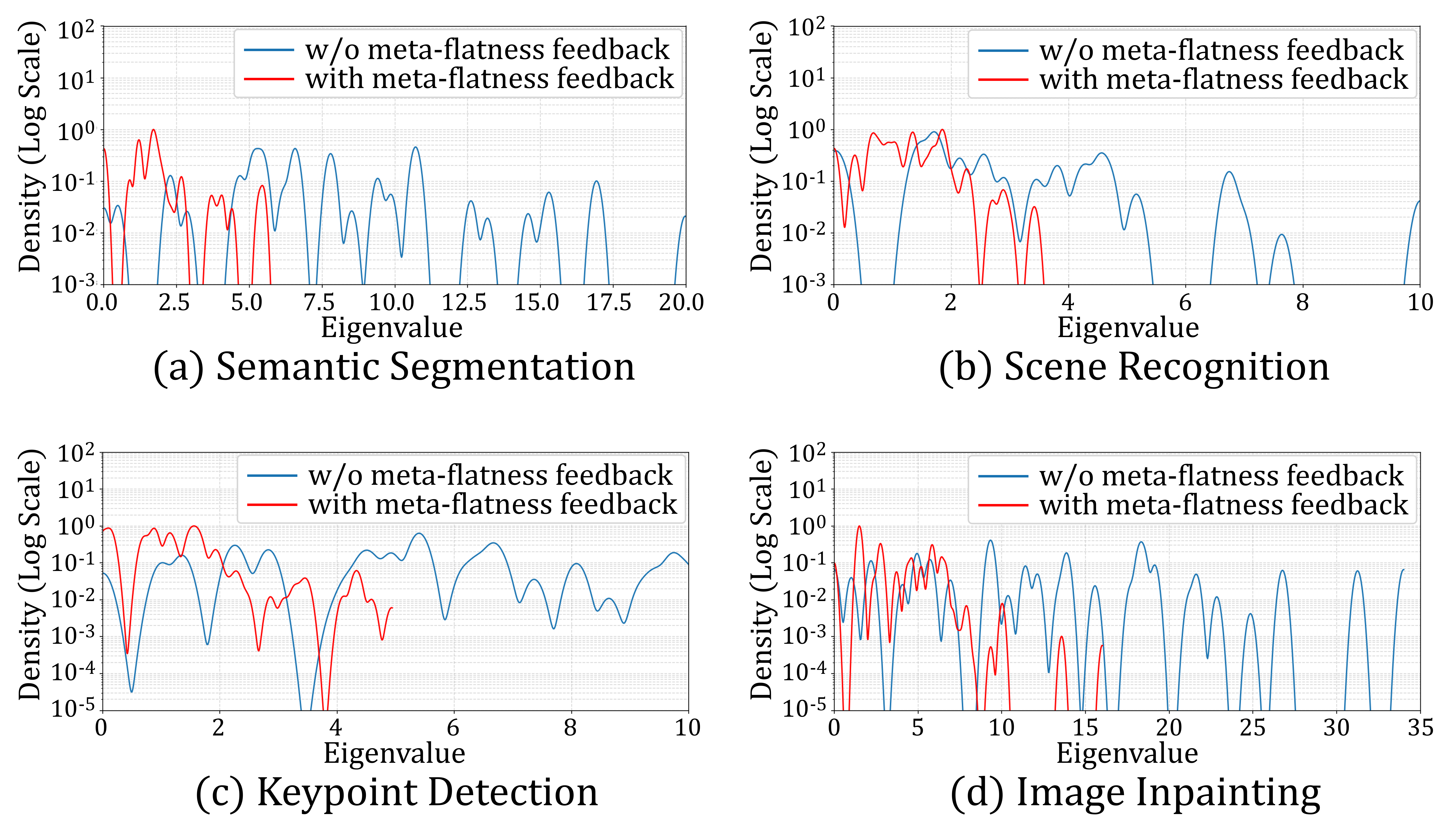}
\caption{Visualization on the Hessian spectra calculate over unseen tasks, on the Taskonomy dataset \cite{zamir2018taskonomy}. Note that obtaining the Hessian spectrum with respect to a certain task does not require the update of the generator. This means that we can formulate the Hessian spectrum over unseen tasks as well.}
\label{fig:visual_2}
\end{figure}

Moreover, note that calculating the Hessian spectrum with respect to a given task does not require updating (training) the unexploitable example generator on that task. This allows us to compute Hessian spectra not only for seen tasks but also for unseen tasks used during evaluation (e.g., the four tasks used to evaluate the generator in Experiment B.1). In Fig.~\ref{fig:visual_2}, we visualize the Hessian spectra for these four unseen tasks on the Taskonomy dataset \cite{zamir2018taskonomy} in the same manner as in Fig.~\ref{fig:visual_1}. As shown, with our meta-flatness feedback, the Hessian spectra on these unseen tasks likewise exhibit peaks that are more concentrated on the left-hand side. This further implies that our method indeed helps converge the unexploitable example generator toward common flat minima in landscape across different tasks, even over unseen tasks.

\subsection{Further Analysis and Evaluation}

\noindent\textbf{Further cross-task evaluation of our framework on image synthesis and 3D reconstruction tasks.}
In our main experiments, on the one hand, in Protocol A, we follow the setting of \cite{sun2024unseg} and evaluate our framework across three tasks (panoptic segmentation, instance segmentation, and object detection) on the COCO dataset. Meanwhile, in Protocol B, we further evaluate our framework across four additional tasks (semantic segmentation, scene recognition, keypoint detection, and image inpainting) on another large-scale multi-task dataset, Taskonomy. To further extend our evaluation, we additionally assess our framework on image synthesis and 3D reconstruction tasks. In specific, we take the generator trained in Experimental Setting B.1 under Protocol B on six interactive segmentation tasks (point-based, box-based, mask-based, point-and-box-based, point-and-mask-based, and box-and-mask-based) and evaluate it on tasks that differ substantially from interactive segmentation, including unconditional image synthesis and multi-view 3D object reconstruction. For unconditional image synthesis, we follow the setup of \cite{ho2020denoising}: using DDPM \cite{ho2020denoising} as the target model, evaluating on the CIFAR-10 dataset \cite{krizhevsky2009learning}, and reporting the FID metric. For multi-view 3D object reconstruction, we follow the setup of \cite{huang20242d}: using 2DGS \cite{huang20242d} as the target framework, evaluating on the DTU dataset \cite{jensen2014large}, and reporting the Chamfer Distance metric. As shown in Tab.~\ref{Tab:ablation_new_task}, across these synthesis and reconstruction tasks, both of which are unseen during the generator’s training process, our framework also significantly outperforms the baseline method UnSeg \cite{sun2024unseg}. This further demonstrates the effectiveness of our framework.

\begin{table}[H]
\caption{Further cross-task evaluation of our framework on image synthesis and 3D reconstruction tasks.}
\centering
\resizebox{0.9\columnwidth}{!}
{
\setlength{\tabcolsep}{2pt}
\begin{tabular}{l|ccccc} \hline

\textbf{Method} &  \makecell{\textbf{Unconditional} \\ \textbf{Image Synthesis} \\ \textbf{(FID$\uparrow$)}} & \makecell{\textbf{Multi-view} \\ \textbf{3D Object Reconstruction} \\ \textbf{(Chamfer distance $\uparrow$)}}\\
\hline
Raw data & 3.17 & 0.80 \\
\hline
Baseline(UnSeg) & 4.21 & 0.93\\
\hline
MCT-UEG & \textbf{58.61} & \textbf{3.13}\\
\hline
\end{tabular}}
\label{Tab:ablation_new_task}
\end{table}

\begin{figure}[t]
\centering
\includegraphics[width=1\columnwidth]{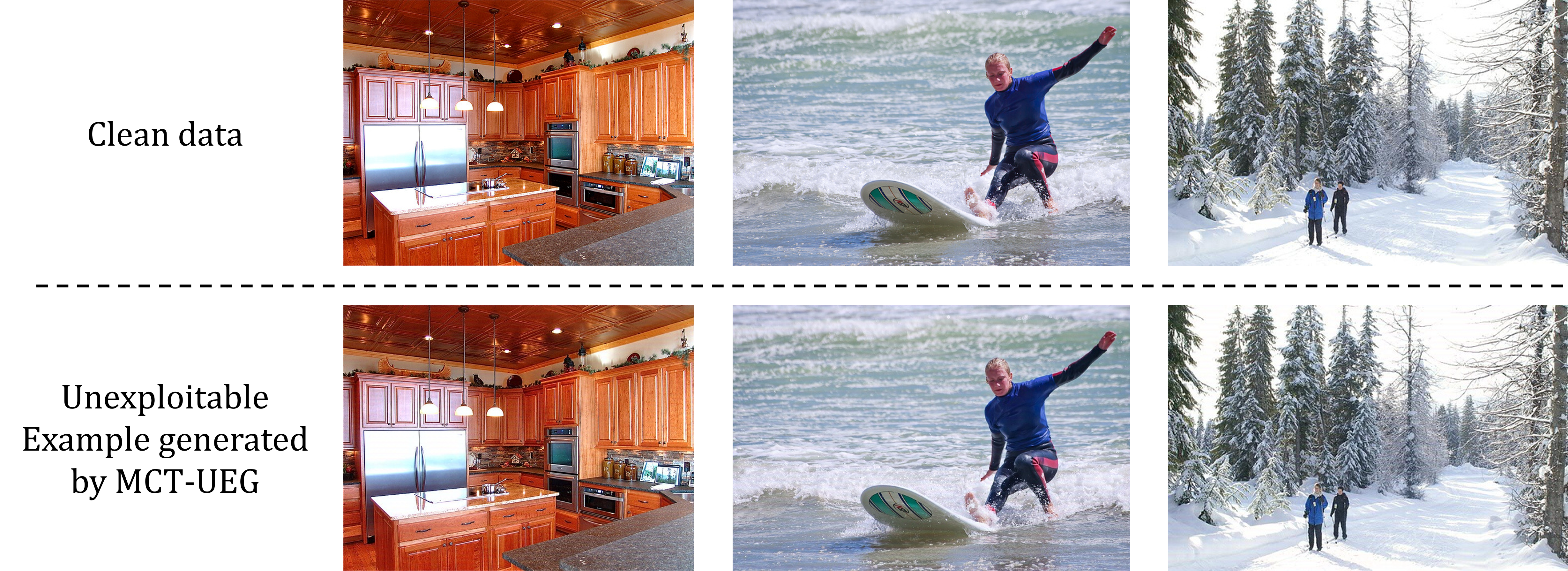}
\caption{Visualization of our generated unexploitable examples. Specifically, we show the clean images on the left-hand side and the corresponding unexploitable examples generated by MCT-UEG on the right-hand side. As shown, when transforming clean images into their cross-task unexploitable versions, our MCT-UEG framework introduces only an imperceptible visual discrepancy. More visualization results are in Supplementary.} 
\label{fig:visual_3}
\end{figure}

\noindent\textbf{Visualization of our generated unexploitable examples.} In this work, we propose the MCT-UEG framework, which aims to make personal images resistant to exploitation across different computer vision tasks by adding imperceptible noise to each image. Beyond demonstrating the cross-task unexploitability of our generated examples through extensive experiments above, we further visualize the unexploitable examples produced by our framework. As shown in Fig.~\ref{fig:visual_3}, when transforming clean personal images into their cross-task unexploitable versions, our framework introduces only an imperceptible visual discrepancy. Consequently, our framework has minimal impact on the normal utility of the images (e.g., sharing on social media).

\section{Conclusion}
\label{sec:conclusion}

In this work, we have proposed a novel unexploitable example generation framework, MCT-UEG. At its core, MCT-UEG features a carefully designed \textit{flat-minima-oriented meta-training and testing} scheme. Leveraging this design, MCT-UEG facilitates the practically effective transformation of personal images into examples that are broadly unexploitable across different computer vision tasks. Extensive experiments have shown the efficacy of our framework in different settings. 

\bibliographystyle{IEEEtran}
{\tiny \bibliography{egbib}}

\begin{IEEEbiography}[{\includegraphics[width=1in,height=1.25in,clip,keepaspectratio]{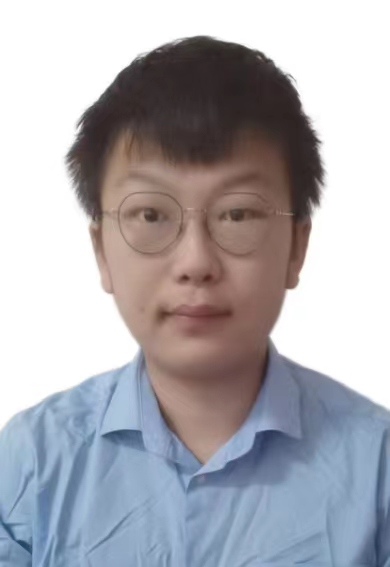}}]{Haoxuan Qu} received the Bachelor of Computing degree in Computer Science and Bachelor of Science degree in Applied Math from National University of Singapore (NUS), Singapore, in 2021. He is currently pursuing the Ph.D degree in School of Computing and Communications in Lancaster University. His research interests are mainly focused on deep learning and computer vision.
\end{IEEEbiography}

\begin{IEEEbiography}[{\includegraphics[width=1in,height=1.25in,clip,keepaspectratio]{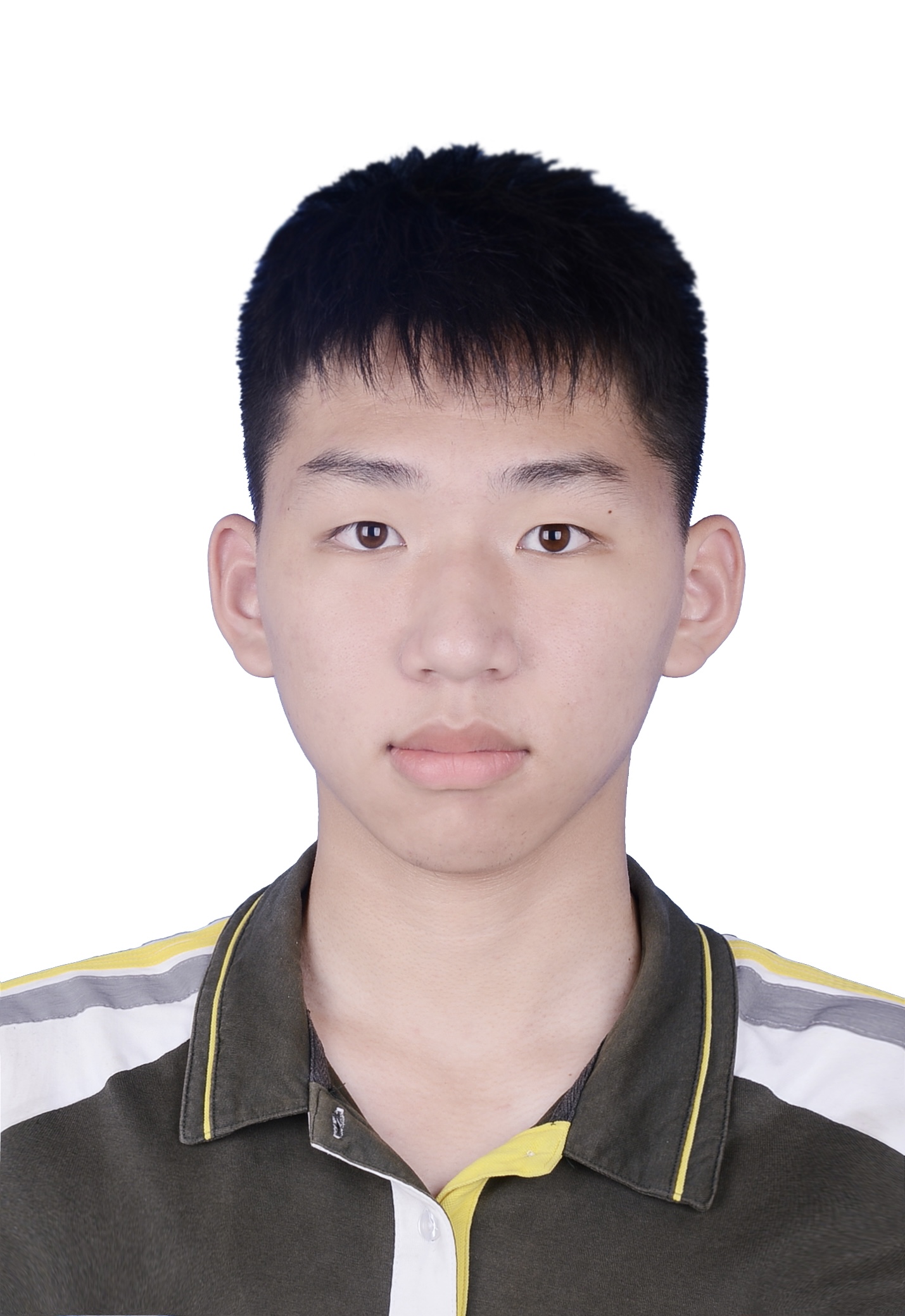}}]{Qiuchi Xiang} received the Bachelor of Engineering degree in Data Science and Big Data Technology from South China Normal University, China, in 2025. He is currently pursuing the Ph.D degree in the School of Computing and Communications at Lancaster University. His research interests are mainly focused on deep learning and computer vision. 
\end{IEEEbiography}

\begin{IEEEbiography}[{\includegraphics[width=1in,height=1.25in,clip,keepaspectratio]{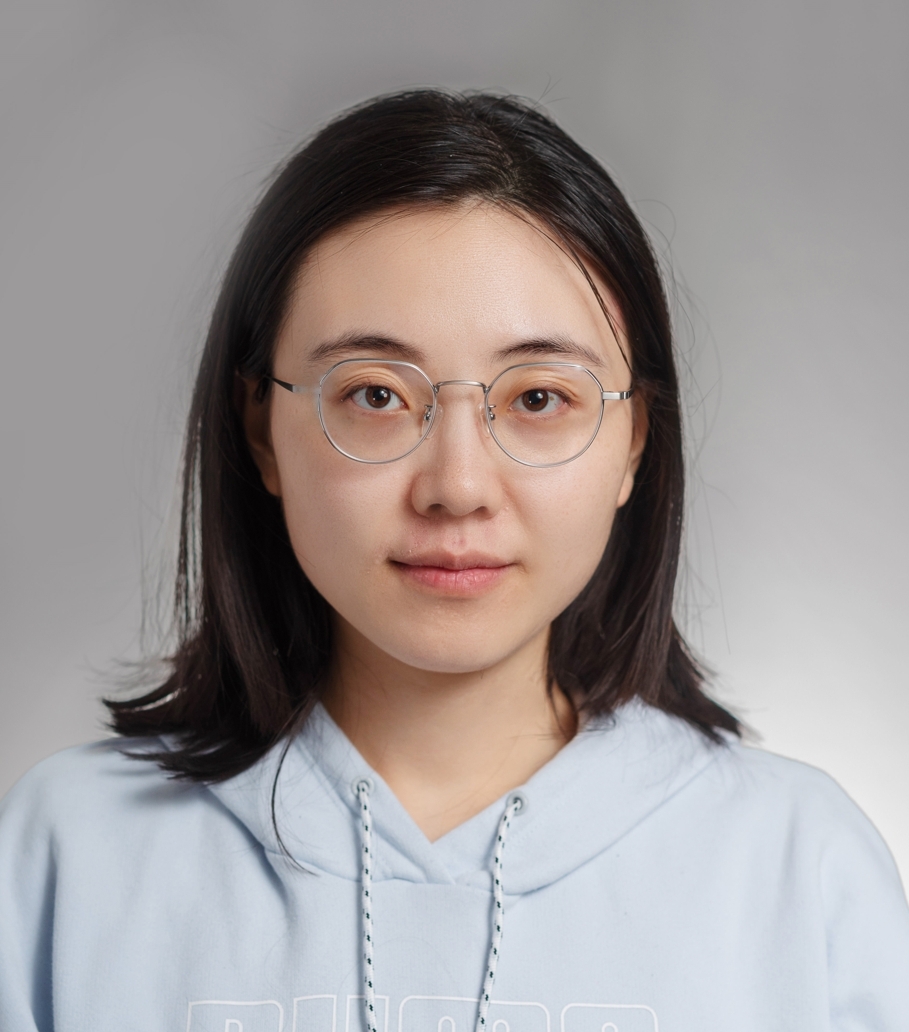}}]{Yujun Cai} is currently a lecturer at the School of Electrical Engineering and Computer Science, the University of Queensland (UQ). Before that she was a research scientist in Meta Inc. She obtained her Ph.D. from Nanyang Technological University in 2021. She served as Area Chair for ICML, WWW, ICCV, IJCAI, WACV etc. She’s also reviewer of IEEE Transactions on Pattern Analysis and Machine Intelligence, International Journal of Computer Vision, IEEE Transactions on Image Processing, IEEE Transactions on Circuits and Systems for Video Technology etc.
\end{IEEEbiography}

\begin{IEEEbiography}[{\includegraphics[width=1in,height=1.25in,clip,keepaspectratio]{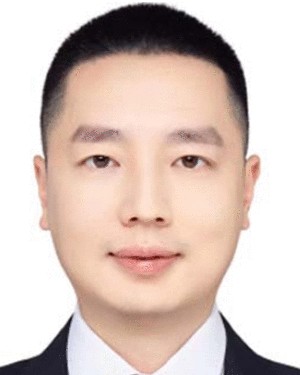}}]{Yirui Wu} (Senior Member, IEEE) received the BS and PhD degrees in computer science and technology from Nanjing University, Nanjing, China, in 2011 and 2016, respectively. He visited Hong Kong University of Science and Technology twice, in 2012 and 2014. He is a young professor with Hohai University, working as a member of the Key Laboratory of Water Big Data Technology of Ministry of Water Resources and the leader of the HHU-Delta Lab. His research interests include computer vision and multimedia understanding.
\end{IEEEbiography}

\begin{IEEEbiography}[{\includegraphics[width=1in,height=1.25in,clip,keepaspectratio]{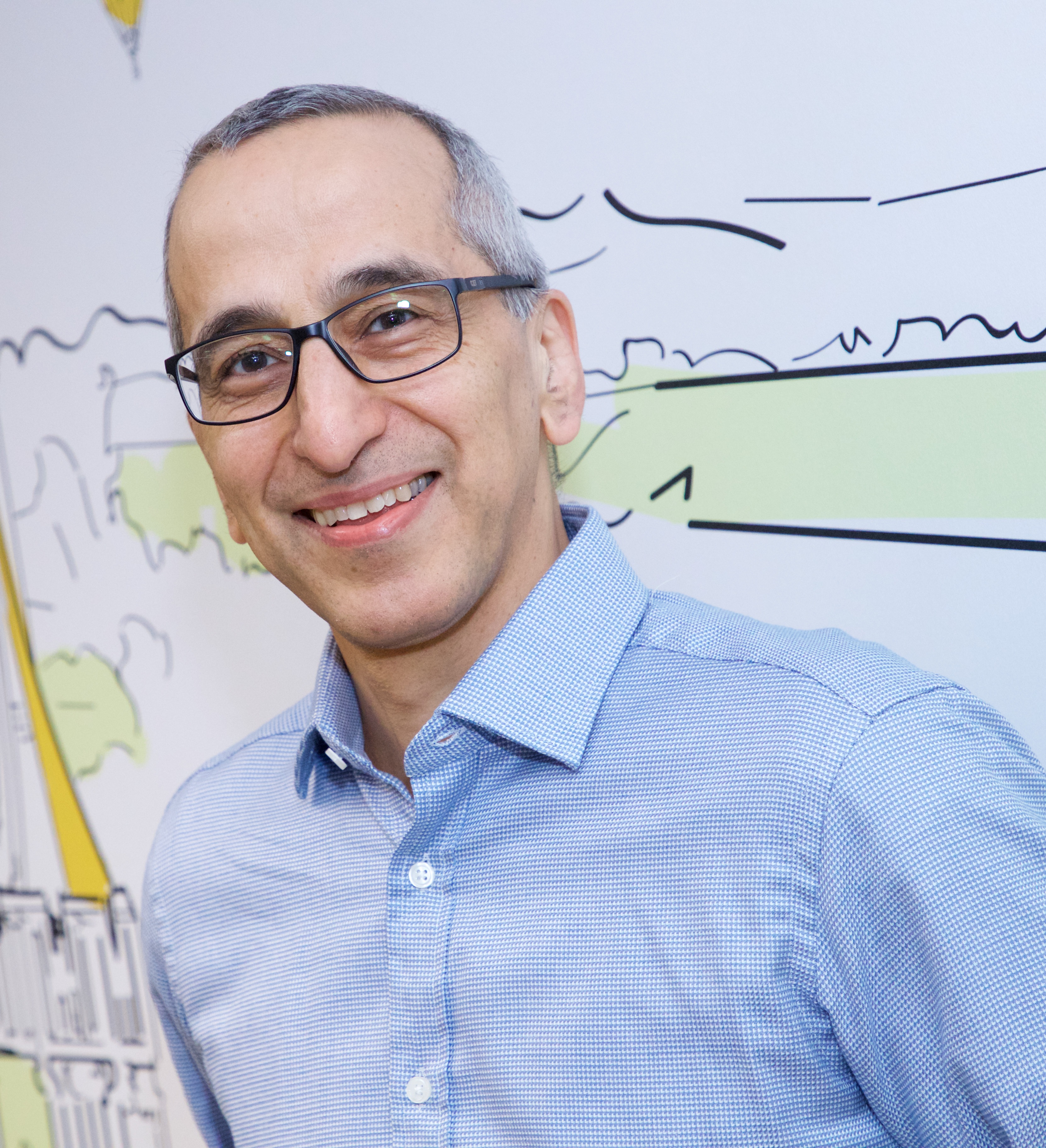}}]{Majid Mirmehdi} is a Professor of Computer Vision in the School of Computer Science at the University of Bristol. His research interests include natural scene analysis, healthcare monitoring using vision and other sensors, and human and animal motion and behaviour understanding. He has more than 250 refereed journal and conference publications in these and other areas. MM is a Fellow of the International Association for Pattern Recognition and a Distinguished Fellow of the British Machine Vision Association. He is Editor-in-Chief of IET Computer Vision journal and an Associate Editor of Pattern Recognition journal. He is a member of the IET and serves on the Executive Committee of the British Machine Vision Association.
\end{IEEEbiography}

\begin{IEEEbiography}[{\includegraphics[width=1in,height=1.25in,clip,keepaspectratio]{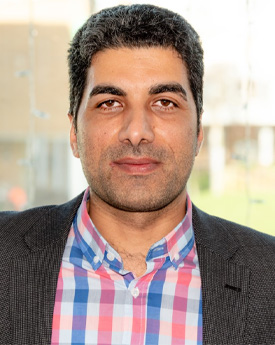}}]{Hossein Rahmani} received the PhD degree from the University of Western Australia, Perth, WA, Australia, in 2016. He is a professor with the School of Computing and Communications, Lancaster University in the U.K. Before that, he was a research fellow with the School of Computer Science and Software Engineering, University of Western Australia. His research interests include computer vision, action recognition, pose estimation, and deep learning. He now serves as an associate editor for IEEE Transactions On Neural Networks And Learning Systems and the Pattern Recognition journal. Meanwhile, he is Program Chair of BMVC 2026. He has served as an Area Chair of CVPR, ECCV, and ICLR, and a Senior Program Committee member of IJCAI.
\end{IEEEbiography}

\begin{IEEEbiography}[{\includegraphics[width=1in,height=1.25in,clip,keepaspectratio]{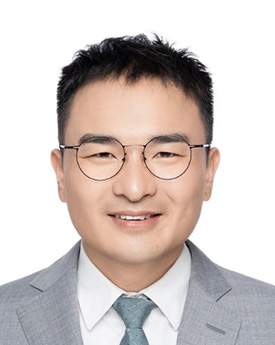}}]{Jun Liu} (Senior Member, IEEE) received the PhD degree from Nanyang Technological University, in 2019. He is a professor and chair in digital health with the School of Computing and Communications, Lancaster University. He was with Singapore University of Technology and Design from 2019 to 2024. He is a Senior Area Editor of IEEE Transactions on Image Processing, and an Associate Editor of IEEE Transactions on Circuits and Systems for Video Technology, IEEE Transactions on Industrial Informatics, IEEE Transactions on Biometrics, Behavior and Identity Science, ACM Computing Surveys, and Pattern Recognition. He is General Chair of BMVC 2026 and Program Chair of BMVC 2025, and has served as an Area Chair of CVPR, ECCV, ICML, NeurIPS, ICLR, AAAI, IJCAI, WACV, and MM. His research interests include computer vision, machine learning and digital health.
\end{IEEEbiography}

\vfill

\end{document}